\documentclass[runningheads]{llncs}
\usepackage{graphicx}
\usepackage{comment}
\usepackage{amsmath,amssymb} % define this before the line numbering.
\usepackage{color}
\usepackage{epsfig}
\usepackage{graphicx}
\usepackage{amsmath}
\usepackage{amssymb}
\usepackage{cite}
\usepackage{multirow}
\usepackage{booktabs}
\usepackage{nth}
\usepackage{bbm}
\usepackage{wrapfig}
\usepackage[ruled,vlined]{algorithm2e}

\usepackage[pagebackref=true,breaklinks=true,colorlinks,bookmarks=false]{hyperref}

\iftrue
\newcommand{\davide}[1]{\textcolor{blue}{\bf [DAVIDE: #1]}}
\newcommand{\andrew}[1]{\textcolor{green}{\bf [ANDREW: #1]}}
\newcommand{\bing}[1]{\textcolor{cyan}{\bf [BING: #1]}}
\newcommand{\joe}[1]{\textcolor{magenta}{\bf [JOE: #1]}}
\newcommand{\todo}[1]{\textcolor{red}{\bf [TODO: #1]}}

\else
\newcommand{\davide}[1]{}
\newcommand{\andrew}[1]{}
\newcommand{\bing}[1]{}
\newcommand{\joe}[1]{}
\newcommand{\todo}[1]{}
\fi

\begin{document}
% \renewcommand\thelinenumber{\color[rgb]{0.2,0.5,0.8}\normalfont\sffamily\scriptsize\arabic{linenumber}\color[rgb]{0,0,0}}
% \renewcommand\makeLineNumber {\hss\thelinenumber\ \hspace{6mm} \rlap{\hskip\textwidth\ \hspace{6.5mm}\thelinenumber}}
% \linenumbers
\pagestyle{headings}
\mainmatter
\def\ECCVSubNumber{1072}  % Insert your submission number here

\title{Multiple Object Tracking with Siamese Track-RCNN} % Replace with your title

% INITIAL SUBMISSION 
\begin{comment}
\titlerunning{ECCV-20 submission ID \ECCVSubNumber} 
\authorrunning{ECCV-20 submission ID \ECCVSubNumber} 
\author{Anonymous ECCV submission}
\institute{Paper ID \ECCVSubNumber}
\end{comment}
%******************

% CAMERA READY SUBMISSION
% \begin{comment}
% \titlerunning{Abbreviated paper title}
% If the paper title is too long for the running head, you can set
% an abbreviated paper title here
%
\author{Bing Shuai \and
Andrew G. Berneshawi\and
Davide Modolo\and
Joseph Tighe
}
\authorrunning{Bing Shuai et al.}
% First names are abbreviated in the running head.
% If there are more than two authors, 'et al.' is used.
%
\institute{Amazon Web Service (AWS) Rekognition \\
\email{\{bshuai, bernea, dmodolo, tighej\}@amazon.com}}
% \end{comment}
%******************
\maketitle

\begin{abstract}

Multi-object tracking systems often consist of a combination of a detector, a short term linker, a re-identification feature extractor and a solver that takes the output from these separate components and makes a final prediction. Differently, this work aims to unify all these in a single tracking system. Towards this, we propose Siamese Track-RCNN, a two stage detect-and-track framework which consists of three functional branches: (1) the detection branch localizes object instances; (2) the Siamese-based track branch estimates the object motion and (3) the object re-identification branch re-activates the previously terminated tracks when they re-emerge.  We test our tracking system on two popular datasets of the MOTChallenge. Siamese Track-RCNN achieves significantly higher results than the state-of-the-art, while also being much more efficient, thanks to its unified design. 
%We present a single network with a shared back bone and three heads that are 

%\todo{WIP: no need to read this} Tracking multiple objects  involves a wide range of video recognition tasks. Thus, the ``Tracking-by-detection''  framework usually trains a hodgepodge of different functional models, which is computationally inefficient during inference.  In this paper, we propose Siamese Track-RCNN, an unified multi-object tracking framework that includes essential components for long-term tracking. In a nutshell,   it is a two stage detect-and-track framework which have three different functional branches: (1), the detection branch detects object instances of interests; (2), the Siamese-based track branch estimates the object motion and (3), the object re-identification re-activates the previously killed tracks when they re-emerge.  
% The online inference of Siamese Track-RCNN requires a single feed-forward pass of the network, therefore it is efficient. Moreover, it is able to produce long, consistent object trajectories. Without bells and whistles, our online Siamese Track-RCNN achieves state-of-the-art tracking performance (MOTA, IDF1) on public MOT Challenge benchmark. Our model outperforms Tracktor++ by a significant 6.1\% MOTA and 6.8\% IDF1 on MOT17, and by 5.3\% MOTA and 6.6\% IDF1 on MOT16.
\end{abstract}

\section{Introduction}
Multi-object tracking (MOT) deals with the problem of localizing and tracking object instances over entire video sequences. Recently, the most successful approaches in the literature are based on the ``tracking-by-detection'' paradigm, which consists of two major components: object detection and association. First, a model localizes all instances of a pre-defined object class in every video frame, and then, it links these together to form object tracks of arbitrary length. 

While object class detection has improved considerably in the last few years~\cite{girshick2015fast, ren2015faster,dai2016r,he2017mask,lin2017feature,dai2017deformable}, association remains a challenging task. Many approaches modelled this as a graph-based optimization problem, where each node is a detection and each edge a possible link~\cite{zhang2008global,berclaz2011multiple,zamir2012gmcp,kim2015multiple,tang2017multiple,henschel2017improvements,ristani2018features,keuper2018motion,sheng2018heterogeneous,xu2019spatial,xu2019train}. These approaches depend on a complex combination of multiple cues, like appearance, re-id, motion estimation and object interactions to solve this association. 
These have helped MOT models gradually improve their performance~\cite{milan2016mot16}, but at the cost of inefficiency and high computation, as each of the employed cues is often its own deep network trained on its own specific dataset. 

Bergmann et al.~\cite{bergmann2019tracking} identified this shortcoming and as a step towards accurate and fast MOT proposed Tracktor, an efficient tracking framework that only relies on a detection module. While Tracktor achieves state-of-the-art performance on the MOTChallenge~\cite{milan2016mot16}, it suffers from two limitations: (i) it does not capture the appearance of the target object during tracking, which causes the model to become uncertain when the object is occluded or there are multiple of the same type of object in close proximity; (ii) it only performs short-term tracking, as it cannot re-identify the target object after it temporally disappears. Bergmann et al. tackled the second limitation by enhancing Tracktor with a re-id network (i.e., Tracktor++), which led to better performance, but at the cost of efficiency, partially defeating the benefits of Tracktor.

In this paper we propose a novel MOT approach: {\it Siamese Track-RCNN}. Differently from the previous works, we address all the aforementioned problems in a single, unified network architecture. Our approach is both efficient and accurate. Our proposed system performs both detection and association in a single forward pass of its network, which consists of a backbone shared among three branches: detection, tracking and re-identification. This design has several advantages: low computational cost, low memory footprint and higher accuracy, as all these branches can benefit from each other. At a high level, the detection branch localizes people entering the field of view, the track branch follows them into the proceeding frames and the re-id branch is responsible for associations across longer periods of time.  

We conduct experiments on the problem of tracking people, which is particularly interesting both for real-world applications, as well as for downstream tasks, like action detection in videos~\cite{zhang2019structured, li2018recurrent, gao2018ctap, choi2012unified}. Our Siamese Track-RCNN achieves state-of-the-art (SOTA) performance on both MOT16 (59.8  MOTA) and MOT17 (59.6 MOTA) of the MOTChallenge. 
% Importantly, it achieves the largest improvement obtained by any method, in recent years, on the MOTChallenge: +5.0 and +5.9 MOTA on MOT16 and MOT17, respectively. \bing{This seems too aggressive.} \andrew{I agree at the least I wouldn't say 'importantly'. Maybe something like 'Given previous year over year gains of a few points we achieve a significant improvement of +xx'} 
Finally, we present an extensive ablation study of Siamese Track-RCNN on a large-scale synthetic dataset: Joint Track Auto (JTA)~\cite{fabbri2018learning}. We show that our model can accurately localize person instances in crowded scenes and consistently track them over long periods of time. 

% To summarize, we make four contributions:
% \begin{enumerate}
%     \vspace{-2mm}
%     \item we propose to unify both detection and association in a single MOT network; \bing{To be fair, this is not the first work to do that.}
%     \item we present a novel MOT architecture called Siamese Track-RNN, which is both efficient and accurate;
%     \item we achieve SOTA results on both the MOT16 and MOT17 sets of the MOTChallenge;
%     \item we demonstrate the merit of the components of the proposed approach on the largest publicly available tracking dataset: JTA. \andrew{Given previous line we already demonstrated the merit on MOT no? How do we differentiate this.}
%\end{enumerate}

\section{Related work}
\subsection{Multi-object tracking (MOT)}
Most works on MOT adopt the ``tracking-by-detection" framework paradigm ~\cite{zhang2008global,berclaz2011multiple,zamir2012gmcp,kim2015multiple,tang2017multiple,henschel2017improvements,ristani2018features,keuper2018motion,sheng2018heterogeneous,xu2019spatial,xu2019train, sadeghian2017tracking, leal2016learning, wang2019exploit, andriyenko2011multi, berclaz2006robust, choi2010multiple, evangelidis2008parametric, fang2018recurrent}, in which detected object instances are associated across time based on their visual coherence and spatial-temporal consistency. 
Some of these works focused on learning new functions to evaluate short-term associations more robustly \cite{ristani2018features, sheng2018heterogeneous, tang2017multiple, xu2019spatial, zhang2008global, tang2017multiple, sadeghian2017tracking, leal2016learning, choi2015near, fang2018recurrent, zhang2008global}. Others, instead, focused on learning how to output more temporally consistent long-term tracks by optimizing locally connected graphs~\cite{zhang2008global,berclaz2011multiple,zamir2012gmcp,kim2015multiple,tang2017multiple,henschel2017improvements,ristani2018features,keuper2018motion,sheng2018heterogeneous,xu2019spatial,xu2019train, wang2019exploit, andriyenko2011multi, berclaz2006robust, evangelidis2008parametric, tang2017multiple}. These are often inefficient, as they employ computationally expensive cues, like object detection~\cite{girshick2015fast, ren2015faster, dai2016r, he2017mask}, optical flow~\cite{dosovitskiy2015flownet, sun2018pwc, tang2017multiple, choi2015near}, re-identification~\cite{hermans2017defense, tang2017multiple, zhou2019deep, tang2017multiple, ristani2018features}. 

Tracktor~\cite{bergmann2019tracking} is the most related to this work, as it also aims at achieving both accurate and fast tracking inference. Towards this, Bergmann et al. proposed to estimate the location of a person in the new frame by regressing his/her position from the previous frame. While Tracktor achieves great results on the MOTChallenge, it does not learn the appearance of people, which is necessary for tracking them through occlusion (e.g., intersections) and for long-term tracking. Instead, we propose a new MOT approach that is more accurate, equally fast and it can deal well with these shortcomings.

%to streamline the multi-object tracking pipeline. Different from previous work, both Siamese Track-RCNN and Tracktor are online multi-object tracker (MOT), and they are efficient. As we demonstrate in our ablation analysis (c.f. section \ref{section:ablation_tracktor}), Tracktor can only perform short-term tracking, whereas our Siamese Track-RCNN can produce consistent and long-term object trajectories.

\subsection{Siamese-based trackers for single-object tracking (SOT)}
Siamese-based trackers~\cite{held2016learning, bertinetto2016fully, li2018high, li2019siamrpn++, tao2016siamese, valmadre2017end, guo2017learning, he2018twofold, zhang2019structured, zhu2018distractor, fan2019siamese, zhang2019deeper} have recently achieved great results in SOT.
% Single object tracking refers to tracking a given object instance (usually in the first frame). It is a challenging task as visual appearance of object instances can change dramatically across temporally proximal frames due to sudden changes of object pose, light conditions, motion blur, etc. In the mean time, the location of objects can shift significantly due to fast motion of obejcts or camera motion. Recently, Siamese-based trackers achieve great success. 
As the name suggests, Siamese approaches operate on pairs of frame. Their goal is to track (by matching) the target object in the first frame in a search region from the second frame. This matching function is usually learned offline on large-scale datasets of image pairs. Among these methods, GOTURN~\cite{held2016learning} inputs the features of both the target object and its corresponding search region to a regression network, which directly estimates the new location of the object.  SiamFC \cite{bertinetto2016fully} solves this matching via correlation operation, where the region with the highest correlation score gives rise to the new object location. SiameseRPN \cite{li2018high}, and later SiamRPN++ \cite{li2019siamrpn++}, adopt an RPN network~\cite{ren2015faster} to explicitly produce the candidate target regions. SiamMask \cite{wang2019fast} further extends these to track object masks. In this paper we are the first, to the best of our knowledge, to propose to use Siamese-based trackers in an end-to-end trainable framework for MOT. Our tracker is inspired by GOTURN, but it is very generic and, in practice, it can integrate any SOT Siamese tracker.

%\subsection{Two-stage R-CNN for tracking}
%Fast and Faster R-CNN \cite{girshick2015fast,ren2015faster} were originally introduced for object class detection in images and Mask-RCNN~\cite{he2017mask} extended these to instance segmentation.  % The first stage of RCNN is a region proposal network (RPN), and the second stage includes different functional components ranging from detection \cite{ren2015faster}\cite{he2017mask} to pixel-wise mask estimation \cite{he2017mask}.
\subsection{Towards a unified network for MOT tracking}
Voigtlaender et al. \cite{voigtlaender2019mots} and Yang et al. \cite{yang2019video} recently proposed to unify detection and re-id for online instance mask tracking. Both of these works built upon Mask-RCNN \cite{he2017mask} and enriched it with a re-identification (re-id) branch. At inference, they associate tracks over time using the instance-specific embedding features outputted by this re-id branch.
In this paper we show that tracking by re-id alone is not robust to short-term changes, especially in the presence of partial occlusions. Instead, we propose to enrich the popular Faster-RCNN~\cite{ren2015faster} object detector (we track bounding boxes, not masks) with both a re-id branch and a tracking branch. We show that these are highly complementary and their combination leads to SOTA MOT performance.

% --------------------------------------------------------- SECTION 3 ---------------------------------------------------------
\section{Siamese Track-RCNN} \label{sec:architecture}

\begin{figure}[t]
  \includegraphics[width=1.0\textwidth]{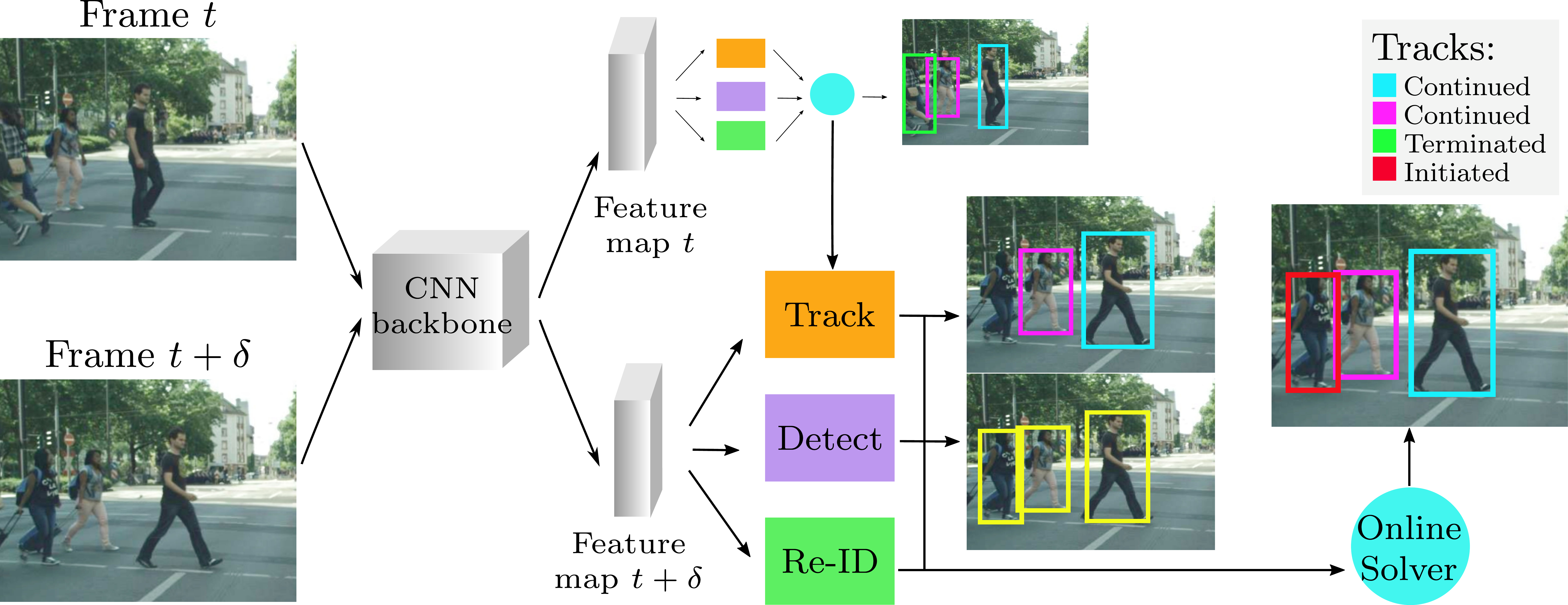}
  \caption{\it Siamese Track-RCNN unifies detection, tracking and re-id in a single network architecture. Importantly, these features share the same backbone, which results in low computation and efficient runtime. }
  \label{figure:siamese_track_rcnn}
\end{figure}

Our tracking system follows an inference pipeline similar to other tracking-by-detection systems, but it does so from cues generated by a single network. We first give a high-level description of our online tracking inference with an eye toward identifying the cues that need to be generated by our approach and then describe the structure of our unified tracking network in detail.

Our Siamese Track-RCNN tracking method operates {\it online}. As such, it tracks people forward in time by only propagating information from the past. It functions by looking at pairs of frames: \{$t$, $t+\delta$\}, where people in frame $t$ are searched for in frame $t+\delta$. If they are found, their track is {\bf continued} with their new bounding box found in frame $t+\delta$.
If a person is no longer visible, as they either walked outside of the frame or are being occluded by other people or objects, then their track is {\bf terminated}. Next, we {\bf localize} people that are visible at time $t+\delta$, but that were not present at time $t$. When a new person is found, they are either assigned to a newly {\bf initiated} track or to a track that was terminated earlier ({\bf reinstated}). 

Differently from the other methods in the literature ~\cite{zhang2008global,berclaz2011multiple,zamir2012gmcp,kim2015multiple,tang2017multiple,henschel2017improvements,ristani2018features,keuper2018motion,sheng2018heterogeneous,xu2019spatial,xu2019train}, we propose the first framework that can perform all these tasks in a single, unified network architecture. Our Siamese Track-RCNN consists of three branches: track, detect and re-identify (fig.~\ref{figure:siamese_track_rcnn}). (i) The track branch is responsible for tracking existing tracks from frame $t$ to frame $t+\delta$. At a high level, it searches for that same person instance at frame $t+\delta$ in a window around the location where they were previously observed at frame $t$. As long as the person is visible in the search region, the track branch should produce a high score $\hat{v}$ as well as the regressed location $\hat{m}$ to continue the track (fig.~\ref{figure:siamese_track_rcnn}, {\color{magenta}magenta}). When the person is not present in this search region the track branch should produce a low score $\hat{v}$ to terminate the track ({\color{green}green}).  (ii) The detection branch is responsible for localizing all people that appear at time $t+\delta$, independently of their status at time $t$ ({\color{yellow}yellow}). (iii) The re-id branch is responsible for generating discriminative embedding features that can help decide whether a newly localized person belongs to a previously terminated track and should be reinstated or requires a new track to be initiated ({\color{red}red}). 

%\joe{Do we need this paragraph?} \bing{This paragraph tells more details of inference, I vote to move it elsewhere, or remove it. The previous paragraph is clear on high-level inference.} \davide{I agree, it no longer fits here.} Finally, we feed the rich information from three components into an online solver that finalizes the ID of the localized tracks. More specifically, it suppresses double predictions for the same person, when localized by both the track and detection branches (using simple non-maximum-suppression~\cite{girshick2015fast}, fig.~\ref{figure:siamese_track_rcnn}, {\color{magenta}magenta}) and compares the features of a newly localized person against the features of tracks terminated a few frames earlier (it has a small memory for caching past features). If these features match, it reinstates the previous tracks, otherwise it initiates a new one ({\color{red}red}). 

With this high level structure in mind, we design our network by starting with the Faster-RCNN architecture~\cite{ren2015faster,he2017mask} and enriching it with two additional branches: track and re-id. In the following sections we present the details of how we structure and train each branch.

\subsection{Detection branch}
Our detection branch is a common Faster-RCNN architecture~\cite{ren2015faster,he2017mask} that consists of a Region Proposal Network (RPN), followed by classification and regression of the generated proposals. 
%
%The RPN loss is a binary cross-entropy loss, 
The classification loss is a multi-class cross entropy loss and the regression loss is a smooth $\ell_1$-loss \cite{girshick2015fast}. Differently from the original \cite{ren2015faster,he2017mask}, our Siamese Track-RCNN takes pairs Faster-RCNN of frames as input instead of a single image. To accommodate for this difference, we average the detection loss $\ell_{detect}$ from both frames while training our detection branch.

\subsection{Track branch} \label{track_branch}

The task of tracking a particular person instance over time is particularly challenging for three reasons: (i) the appearance of the person can change rapidly over time due to pose changes, occlusion and motion blur; (ii) the location of a person can change quickly due to fast object and camera motion; (iii) tracks of different people can intersect when the individuals cross each other. 
To overcome the first two limitations, we propose a Siamese-based solution that is robust to appearance changes and fast motion. 
For the third, instead, we rely on the re-id branch to help distinguish the two people that are intersecting.
%For people that is occluded due to track intersection, we rely on the re-id branch to associate the split tracks.
% 

Similar to Siamese-based single object tracker in the literature (SOT)~\cite{held2016learning,li2018high,zhu2018distractor,li2019siamrpn++,wang2019fast}, given two frames $t$ and $t+\delta$, we model this problem as matching (searching) a target template from $t$ within a larger contextual region in $t+\delta$.
More specifically, given an ongoing track $\mathcal{T}$ of a person bounding box defined as $target^t=[x^t, y^t, w^t, h^t]$ at time $t$, our branch attempts to localize the person in $t+\delta$ by searching around its previous location at time $t$. 
We obtain our search region by enlarging the original detection $target^t$ by a factor $r$: $search^{t+\delta}=[x^t, y^t, w^t*r, h^t*r]$. Then, we extract the target template feature map $\mathbf{f}_{target^t}$ and  that of its corresponding search region: $\mathbf{f}_{search^{t+\delta}}$. Finally, we predict the status of the person at time $t+\delta$ as follow:
\begin{equation}
    (\hat{v}, \hat{m}) = \mathbb{\phi}(\mathbf{f}_{target^t}, \mathbf{f}_{search^{t+\delta}})
    \label{equation: siamese_track}
\end{equation}
 
 where $\hat{m}$ is the estimated object motion parameterized as $[\frac{x^{t+\delta} - x^t }{w^t}, \ \frac{y^{t+\delta}-y^t}{h^t}$   $\text{log}\frac{w^{t+\delta}}{w^t}], \ \text{log}\frac{h^{t+\delta}}{h^t}]$, $\hat{v}$ indicates whether the person is visible or not at time $t+\delta$, which is essential to ensure that a track does not continue if its person is no longer in view; and $\phi$ is a fully connected network inspired by GOTURN~\cite{held2016learning}. In order for this to work, we flatten $\mathbf{f}_{target^t}$ and $\mathbf{f}_{search^{t+\delta}}$ and concatenate them before inputting them into $\phi$. We chose this specific network for $\phi$ because it is accurate and extremely efficient, but note any Siamese-based SOT~\cite{held2016learning,li2018high,zhu2018distractor,li2019siamrpn++,wang2019fast} can be employed to parameterize $\phi$. This is especially appealing, as our model can directly benefit from future advances in SOT. \\
 
\noindent {\bf Training.} We train our network $\phi$ using a combination of two losses: one over the estimated motion $\hat{m}$ and one over the visibility value $\hat{v}$:
% \begin{equation}
%     \ell_{track} = \ell_{cls}(\hat{v}, v^{gt}) + \mathbbm{1}[v] \ell_{motion}(\hat{p}, p^{gt})
% \end{equation}

\begin{equation}
    \ell_{track} = 
        \begin{cases} 
             \ell_{cls}(\hat{v}, v^{gt}) +  \mathbbm{1}[v^{gt}] \ell_{motion}(\hat{m}, m^{gt}) & \text{if } target^t\in \mathcal{P}^+ \\ \\
             \ell_{cls}(\hat{v}, 0) & \text{if } target^t\in \mathcal{N}^-\\
        \end{cases}
\end{equation}
where $\mathbbm{1}$ is the indicator function ($m^{gt}$ is not defined if the target is not visibile at time $t+\delta$), $\ell_{cls}$ denotes the binary cross entropy loss, $\ell_{motion}$ denotes the commonly used smooth $\ell_1$ loss for regression, $ \mathcal{P}^+$ is the set of positive samples and $ \mathcal{N}^-$ the set of negative ones. Note how in this formulation the ground truth visibly value $v^{gt}$ operates slightly differently depending on the type of sample considered: if $target^t$ is a positive bounding box containing a person, then it indicates whether that person is ($v^{gt}=1$) or is not ($v^{gt} = 0$) visible in frame $t+\delta$; on the other hand, if $target^t$ is a negative bounding box (e.g., a false positive from the detection branch), it is always 0 and the loss encourages the network to immediately stop tracking.
%whatever instance is in $target^t$, even though this may actually be visible in frame $t+\delta$.
In this latter case we also avoid computing any motion estimation, as $m^{gt}$ is not defined for negatives.  

In addition to training on ground truth people and random negatives, we also train on proposals generated by the RPN network. This helps the track branch becoming  more robust to noise in the target location at time $t$. For this, we consider as positive any proposal that has an intersection-over-union (IoU) of at least 0.5 with a ground truth bounding box. Given a pair of frames, we compute this loss between every proposal in frame $t$ and its corresponding region in frame $t+\delta$ and then average over all of them.
Finally, as an additional data augmentation we also compute the inverse tracking from $t+\delta$ to $t$, as this comes almost for free, given that the majority of the computation is spent obtaining the frames' feature vectors from the CNN backbone (fig.~\ref{figure:siamese_track_rcnn}).

\subsection{Re-id branch}
The track branch presented in the previous section only performs short-term tracking, as it can only continue tracks while they are visible. When heavy occlusion happens, we reinstate tracks when they become visible again by employing a re-identification branch that produces an instance-discriminatory embedding for every person instance. \\ %This embedding is then used to compare a new person's detection with those of terminated tracks. \\%If the new person matches (i.e., has similar embedding) any of these, it means that the person was previously being tracked, but terminated because occluded or, more generically, out of view. In this case, its track gets reinstated instead of initiating a new track id. \\

\noindent {\bf Training.} We model our re-id branch as a fully-connected network and train it with a triplet loss~\cite{hermans2017defense}:
\begin{equation}
    \ell_{reid} = \max(0, \max_{q \in \mathcal{P}^p} d(p, q) - \min_{n \in \mathcal{N}^p} d(p, n) + \alpha)
\end{equation}
where $p$ is a reference positive proposal from any of the current pair of frames $\{t, t+\delta\}$, $\mathcal{P}^p$ is the set of positive proposals from the same video of $p$, of the same person (just not in $t$ or $t+\delta$) and $\mathcal{N}^p$ is the set of proposals containing any person except $p$. Finally, $\alpha$ is the distance margin, by which the distance between $p$ and its most similar negative has to be larger than that of $p$ and its least similar positive. 

\subsection{End-to-end training}
In our Siamese Track-RCNN, these three components share the same CNN backbone and are trained together in an end-to-end fashion. This design has several advantages: low computational cost, low memory footprint and potentially higher accuracy, as all these components can benefit from each other. All these are essential for online tracking in real-world applications.
We train our model by simply summing the losses of these components: $\ell = \ell_{detect} +\ell_{track} + \ell_{reid}$.

\section{Implementation details}\label{sec:impl_details}

\paragraph{Network.} As our CNN backbone we use a ResNet-101~\cite{he2016deep} with feature pyramid network~\cite{lin2017feature} and deformable convolutions~\cite{dai2017deformable} (ResNet-101-DCN). The detection branch has the same structure of the popular Faster-RCNN~\cite{ren2015faster}, while the track and re-id branches consist of two fully connnected layers of 1024 and 512 features, respectively. The re-id branch outputs an embedding of 128 features. We follow \cite{held2016learning} and set the search extension ratio $r = 2$, practically doubling the target size, and we set the distance margin $\alpha$ empirically to 0.2. Finally, we use a ROI Align layer~\cite{he2017mask} and pool feature maps of $7\times7$ from frame regions. In the Track branch, we extract these features from both the target bounding box at time $t$ and the enlarged search area at time $t+\delta$, even thought the latter is twice as large. In other experiments we observed that enlarging the feature maps of the search area does not bring any improvement in performance. 
% As previously observed in GOTURN, enlarging the feature map size for the larger search area does not bring any meaningful improvement in performance. \bing{I will check the reference}

\paragraph{Training.} 
We pre-train the backbone and the detection branch for object class detection on the popular MS COCO dataset~\cite{lin2014microsoft} and then finetune the network with all the branches on tracking videos. Pre-training on MS COCO is mostly necessary when the number of training videos available is limited, like in the the case of the MOTChallenge dataset that we present and experiment on in sec.~\ref{sec:mot}. During finetuning we freeze all the batch norm layers \cite{ioffe2015batch}. 
We use stochastic gradient descent to optimize our Siamese Track-RCNN, which is trained for a total of 15k iterations. We start training with a learning rate of 0.02 and decrease it by factor 10 after 10K and 12.5K iterations, respectively. We use a fixed weight decay of $10^{-4}$ and a batch size of 40 frame pairs. 
Finally, we augment our training data of pairs by sampling them randomly within a 1 second temporal window, which is equivalent to setting $\delta = 30$ frames for 30fps videos (range: $[t, t+30]$).

\paragraph{Inference.} At inference we instead set $\delta = 1$ frame, as we aim to keep computation low and run Siamese Track-RCNN as a sliding window. We feed the rich outputs of the three branches into an online solver that finalizes the ID of the localized people.
Specifically, given a set of person localizations, our solver first merges those shared by both the detection and track branches (i.e., those that have intersection-over-union (IoU) $> 0.3$) and then terminates tracks that have a visibility score lower than 0.3 ($\hat{v} < 0.3$). 
Furthermore, it reinstates a previously terminated track when its embedding features are very similar to those of a newly localized person. In practice, the solver postpones this decision to after the new localized person has been tracked for a few frames. Then, it computes the average $\ell_2$ embedding distance between the 5 most similar bounding boxes from the new and old tracks. If this value is less than 0.5, the track gets reinstated. Otherwise, a new track is initiated. 
To enable this feature comparison, our solver maintains and updates a small buffer that caches the embeddings of the terminated and ongoing tracks. We set the size of this buffer to 30 seconds, which offers a good trade-off between low memory consumption and enough temporal information.
%Finally, despite these choices, we have empirically observed that our Siamese Track-RCNN is robust to values in the range $0.3$ to $0.5$ and any choice achieves similar performance. 

\begin{table}[t]
	\begin{center}
	  \resizebox{\textwidth}{!}{%
		\begin{tabular}{l c @{\hskip 0.5em} | c c @{\hskip 0.5em} c @{\hskip 0.5em} c @{\hskip 0.5em} c @{\hskip 0.5em} c c }
			\toprule
			Method & Year & MOTA $\uparrow$ & IDF1 $\uparrow$ & MT $\uparrow$ & ML $\downarrow$ & FP $\downarrow$ & FN $\downarrow$ & IDsw $\downarrow$ \\
			\midrule
			Siamese Track-RCNN & 2020 & \textbf{59.6} & \textbf{60.1} & \textbf{23.9}\% & \textbf{33.9}\% & 15532 & \textbf{210519} & 2068 \\
			\midrule
			DeepMOT-Tracktor \cite{xu2019train} & 2019 & 53.7 & 53.8 & 19.4\% & 36.6\% & \textbf{11731} & 247447 & 1947 \\
			Tracktor++ \cite{bergmann2019tracking} & 2019 & 53.5 & 52.3 & 19.5\% & 36.6\% & 12201 & 248047 & 2072 \\
			DeepMOT-SiamRPN \cite{xu2019train} & 2019 & 52.1 & 47.7 & 16.7\% & 41.7\% & 12132 & 255743 & 2271 \\
			eHAF \cite{sheng2018heterogeneous} & 2018 & 51.8 & 54.7 & 23.4\% & 37.9\% & 33212 & 248047 & \textbf{1834} \\
			FWT \cite{henschel2017improvements} & 2017 & 51.3 & 47.6 & 21.4\% & 35.2\% & 24101 & 247921 & 2648 \\
			jCC \cite{keuper2018motion} & 2018 & 51.2 & 54.5 & 20.9\% & 37.0\% & 25937 & 247822 & 1802 \\
			STRN \cite{xu2019spatial} & 2019 & 50.9 & 56.5 & 20.1\% & 37.0\% & 27532 & 246924 & 2593 \\
			MOTDT17 \cite{chen2018real} & 2018 & 50.9 & 52.7 & 17.5\% & 35.7\% & 24069 & 250768 & 2474 \\
			MHT\_DAM \cite{kim2015multiple} & 2015 & 50.7 & 47.2 & 20.8\% & 36.9\% & 22875 & 252889 & 2314 \\
			\bottomrule
		\end{tabular}}
	\end{center}
	\vspace{-2mm}
	\caption{\it Results on MOT17 test set using the provided public detections.  \vspace{-3mm}}
	\label{table:mot17_results}
\end{table}

\section{Results on MOTChallenge} \label{sec:mot}

\paragraph{Dataset.} 
MOTChallenge~\cite{milan2016mot16} is the most widely-adopted multi-person tracking benchmark. It consists of 7 training and 7 test videos that capture scenes from different camera angles. MOT is challenging, as it contains several occluded or truncated individuals and a large number of people to track. We experiment with two tracking benchmarks in the MOTChallenge: MOT16 and MOT17. The official tracking task consists of correctly linking (tracking) a set of provided per-frame person detections over the whole video. Both sets contain exactly the same videos, but they differ in the provided detections: MOT16 provides DPM~\cite{dpm} detections, while the MOT17 also provides them from Faster R-CNN~\cite{ren2015faster} and SDP~\cite{yang2016exploit}.

\paragraph{Evaluation and metrics.}
We obtain our MOTChallenge results on the test videos (whose annotations are purposely kept private) by submitting our model's predictions to the MOT evaluation server\footnote{\url{https://motchallenge.net/}}. We follow the benchmark's guidelines and  report several metrics, mainly: MOTA (Multiple Object Tracking Accuracy), IDF1 (ID F1 score), MT (Mostly tracked targets), ML (Mostly lost targets), FP (False Positives), FN (False Negatives) and IDsw (ID switches). Among these MOTA is considered the main metric for comparison, while the others are helpful to understand the kind of mistakes a tracker makes. We refer the reader to \cite{milan2016mot16} for more detailed definitions of these metrics.

\paragraph{Experimental settings.} 
We train our Siamese Track-RCNN on the provided training videos, but exclude the detections with visibility score lower than 5\% (this score is provided by MOTChallenge). For a fair comparison with the literature, we present our results using the public detections released by MOTChallenge. 
% To do so, at inference we disable the RPN network of Siamese Track-RCNN and instead input the provided, public detections. 

\begin{table}[t]
    \footnotesize
	\begin{center}
	\resizebox{\textwidth}{!}{%
		\begin{tabular}{l c @{\hskip 0.5em} | c c @{\hskip 0.5em} c @{\hskip 0.5em} c @{\hskip 0.5em} c @{\hskip 0.5em} c c }
			\toprule
			Method & Year & MOTA $\uparrow$ & IDF1 $\uparrow$ & MT $\uparrow$ & ML $\downarrow$ & FP $\downarrow$ & FN $\downarrow$ & IDsw $\downarrow$ \\
			\midrule
			Siamese Track-RCNN & 2020 & \textbf{59.8} & \textbf{60.8} & \textbf{22.0}\% & \textbf{34.5}\% & 4389 & \textbf{68376} & 556 \\
			\midrule
			DeepMOT-Tracktor \cite{xu2019train} & 2019 & 54.8 & 53.4 & 19.1\% & 37.0\% & \textbf{2955} & 78765 & 645 \\
			Tracktor++ \cite{bergmann2019tracking} & 2019  & 54.4 & 52.5 & 19.0\% & 36.9\% & 3280 & 79149 & 682 \\
			DeepMOT-SiamRPN  \cite{xu2019train} & 2019  & 51.8 & 45.5 & 16.1\% & 45.1\% & 3576 & 83699 & 641 \\
			HCC \cite{ma2018customized} & 2018 & 49.3 & 50.7 & 17.8\% & 39.9\% & 5333 &  86795 & \textbf{391} \\
			LMP \cite{tang2017multiple} & 2017 & 48.8 & 51.3 & 18.2\% & 40.1\% & 6654 &  86245 & 481 \\
            STRN \cite{xu2019spatial} & 2019 & 48.5 & 53.9 & 17.0\% & 34.9\% & 9038 & 84178 & 747 \\
			GCRA \cite{ma2018trajectory} & 2018 & 48.2 & 48.6 & 12.9\% & 41.1\% & 5104 &  88586 & 821 \\
			FWT \cite{henschel2017improvements} & 2017 & 47.8 & 44.6 & 19.1\% & 38.2\% & 8886 &  85487 & 852 \\
			MOTDT \cite{chen2018real} & 2018 & 47.6 & 50.9 & 15.2\% & 38.3\% & 9253 &  85431 & 792 \\
			\bottomrule
		\end{tabular}}
	\end{center}
	\vspace{-2mm}
	\caption{\it Results on MOT16 test set using the provided public detections. \vspace{-3mm}}
	\label{table:mot16_results}
\end{table}

\begin{wrapfigure}{r}{0.4\textwidth}
    \centering
    \vspace{-7mm}
    \includegraphics[width=0.4\textwidth]{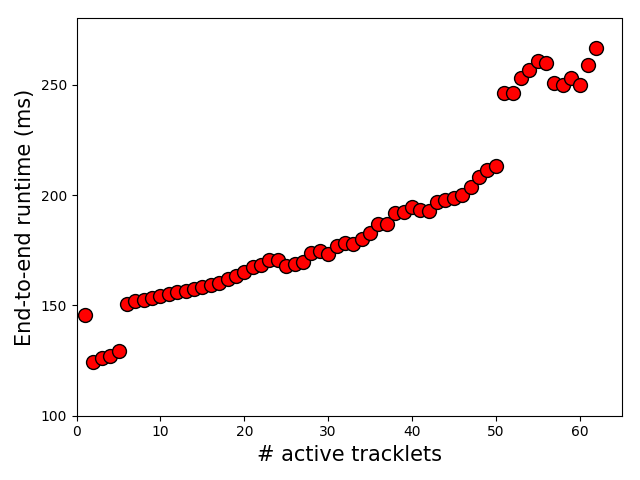}
    \vspace{-7mm}
    \caption{\it End-to-end runtime of our model on the MOT17 videos. \vspace{-5mm}}
    \label{fig:runtime}
\end{wrapfigure}
\paragraph{Results.} We present our results on MOT17 and MOT16 in tables~\ref{table:mot17_results} and \ref{table:mot16_results} respectively. Our Siamese Track-RCNN achieves state-of-the-art (SOTA) MOTA in both datasets. Importantly, the improvement over the previous best method (DeepMOT-Tracktor, which shares a similar backbone as our Siamese Track-RCNN) is considerable: +5.9 on MOT17 and +5.0 on MOT16.

\paragraph{End-to-end runtime evaluation.} Siamese R-CNN does not increase computation significantly compared to its base Faster-RCNN object detector, as it only adds two new branches, which share the same low level computation. In Fig.~\ref{fig:runtime} we show how our model's runtime with respect to the number of ongoing tracks. We test it with a ResNet-101-DCN with FPN backbone, on a single Tesla V100 GPU, on the videos of MOT17 resized to $1080 \times 1920$ pixels. Our runtime's increase is rather sublinear: our model runs at 6 fps when the number of people to track per-frame is at around 20, at 5 fps around 40 and 4 fps around 60.  In terms of end-to-end throughput, this is considerably more efficient than most of the methods in table~\ref{table:mot17_results}, including the previous SOTA Tracktor++.

%We run the model on a single Tesla V100 GPU, and, in Fig. \ref{fig:runtime}, we present the end-to-end average runtime of our Siamese RCNN (with ResNet-101) w.r.t the number of active tracks per frame. Our un-optimized python implementation has a end-to-end throughput of 5 FPS for high-resolution ( $1080 \times 1920$ ) videos even if they include an average 40 persons / frame.  As shown in Fig.\ref{fig:runtime}, the processing time is highly correlated to the number of people per frame, and this is due to the fact that the three branches does not share a significant amount of computation. To further increase the model efficiency, we can optimize the architeture of Siamese Track-RCNN similar to the spirit of R-FCN \cite{dai2016r}.
% It's also important to note that our implementation can be optimized by parallel implementation of all other functional components only except the track branch.

\section{Ablation analysis}
In this section, we carry out an ablation study on Siamese Track-RCNN and its components.

\subsection{JTA dataset} 
% Given the fact that MOT Challenge benchmark has very few training videos and Siamese Track R-CNN is a high-capcity network, MOT is not a great candidate for ablation analysis \footnote{Moreover, it restricts the maximum submission times per tracker to be 4, thus it discourages parameter tuning on MOT test data.}.   
In sec.~\ref{sec:mot} we presented our results on MOTChallenge. While good for comparing the performance of a model against the literature, the dataset is not very appropriate for an ablation study: it does not contain any validation videos and has only 7 training samples, which are too few to further split into train and val. Sadly, there does not exist any publicly available, large scale, real-world multi-person tracking dataset on which we can evaluate.%, which is probably the biggest obstacle today in multi-person tracking research.   

\begin{figure}[t]
    \centering
    \includegraphics[width=1.0\textwidth]{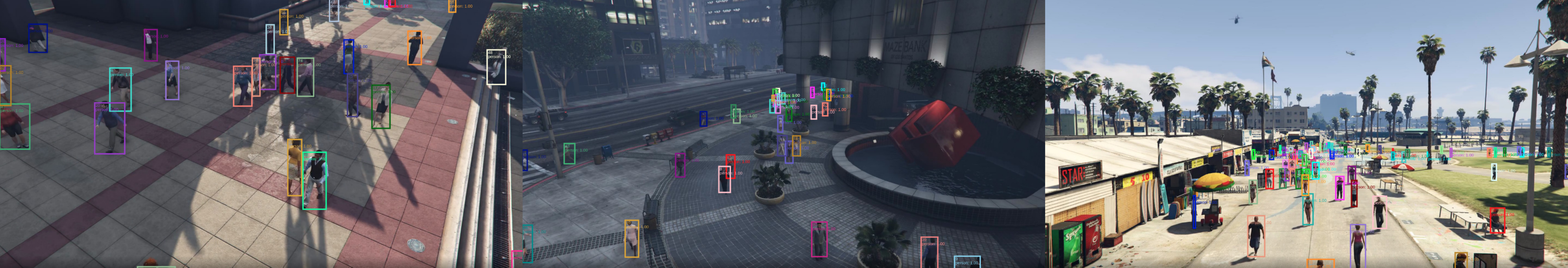}
    \caption{\it JTA dataset: examples of frames and annotations.}
    \label{figure:jta}
\end{figure}

Given the lack of real-world data, we carry out our ablation study on a synthetic dataset: Joint Track Auto (JTA)~\cite{fabbri2018learning}. JTA is a dataset annotated for human pose estimation and tracking and its videos are auto generated using the Grand Theft Auto (GTA) game engine. The dataset contains 512 videos of 500k frames and with almost 10M accurately annotated (since they are machine generated) human poses. Its videos are challenging (fig.~\ref{figure:jta}), as they contain a lot of people, often crossing/occluding each other, and appearing at considerably different scales. These videos are split into train (256), validation (128) and test (128) and in our ablation study we train on ``train'' and evaluate on ``validation''. 

Since the dataset is not annotated with person bounding boxes, we adapt the human pose estimation annotations as follows. First, we take the 22 body joints of each person and fit a tight bounding box around them. Then, we enlarge the bounding box by 5\% in each direction to capture the whole extent of a person (as the joints do not lie on the boundaries). This procedure is very similar to the one employed to generate bounding boxes for PoseTrack 2018~\cite{andriluka2018posetrack}. Finally, we disregard the bounding boxes that are too small ($<25 \times 50$ pixels), that are too far away from the camera ($>25$ meters) and that contain less than 50\% of the body joints. As we downsample all the high-resolution 1080p videos to have a short side of $900$, these small and hard bounding boxes become almost untrackable.

\subsection{Evaluation metric: TrackAP} 
\begin{figure}[t]
  \centering
  \includegraphics[width=1.0\textwidth]{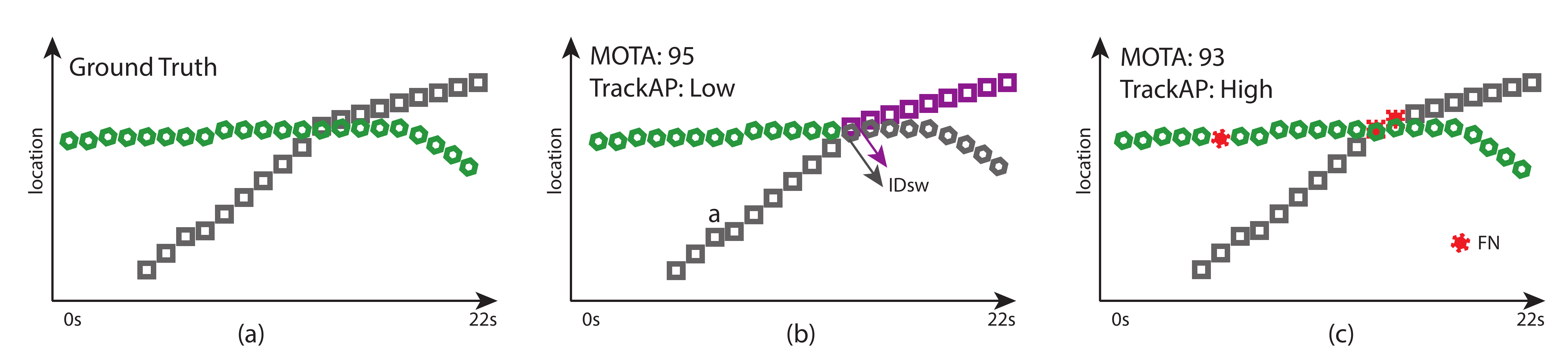}
  \vspace{-4mm}
  \caption{\it While MOTA can be deceivingly high when the detection performance is very good, TrackAP better represents the overall temporal consistency of the tracks. \vspace{-4mm}}
 \label{figure:mota_ap}
\end{figure}

In sec.~\ref{sec:mot} we evaluated on the metrics defined by MOTChallenge. Among those that we reported, MOTA is considered the most important and is often the only metric highlighted. MOTA is indeed a good metric for tracking, but it over-emphasizes the accuracy of the detections of a track, rather than the temporal consistency of the track itself. For example, for MOTA, localizing all people correctly (fig.~\ref{figure:mota_ap}{\color{red}b}) is more important (i.e., it yields better results) than assigning the correct tracking ID to every localized person (fig.~\ref{figure:mota_ap}{\color{red}c}). We argue that this is due to two factors: (i) MOTA weights equally identity switches and false negatives and (ii) false negatives are far more frequent than switches (e.g., in table~\ref{table:mot17_results} our model makes 68376 FN, but only 556 IDsw).

Having temporally consistent tracks is however relevant to many real-world applications, like surveillance and sport, where tracking the person of interest over the whole video sequence is far more important than localizing him precisely. 
In order to better understand the temporal coherence of our model's predictions, we propose to use a second evaluation metric that heavily penalizes id switches. This metric is called {\it TrackAP} (Track Average Precision). It was recently introduced for video instance segmentation~\cite{yang2019video} and it is computed similarly to the object AP for object class detection~\cite{everingham2015pascal,lin2014microsoft}, but with IoU defined over tracks rather than object instances. Specifically, the IoU between a predicted track and its corresponding ground truth equals the average IoU between their bounding boxes: $\text{IoU}_{track}(\hat{\mathcal{T}},\mathcal{T}^{gt}) = \sum_{t=1}^T \text{IoU}_{det}(\hat{bb}_t,bb_t^{gt})$,
% \begin{equation}
%     \text{IoU}(\hat{\mathcal{T}},\mathcal{T}^{gt}) = \frac{\sum_{t=1}^T |\hat{bb}_t \cap bb_t^{gt}|}{\sum_{t=1}^T | \hat{bb}_t \cup bb_t^{gt}|}
% \end{equation}
where $\hat{bb}_t$ is the detection at time $t$ for the predicted track $\hat{\mathcal{T}}$ and $bb_t^{gt}$ is its corresponding ground truth bounding box. If no bounding box is predicted for $bb_t^{gt}$, $\text{IoU}_{det}$ for time $t$ is set to 0.
As for object class detection~\cite{lin2014microsoft}, we threshold $\text{IoU}_{track}$ at 50 and 75 and report AP$_{50}$ and AP$_{75}$.

\subsection{Results on JTA}
\label{section:ablation_tracktor}
We experiment on JTA using three different methods. For an apples to apples comparison, we train them with the same underlying detection module: Faster-RCNN with a ``ResNet50-DCN with FPN'' backbone. At inference, people are tracked using: IoU overlap in Deepsort~\cite{Wojke2017simple}, bounding box regression in Tracktor~\cite{bergmann2019tracking} and the Siamese Track branch of sec.~\ref{track_branch} in the case of our Siamese Track-RCNN. The results are presented in table~\ref{table:results_jta} and reveal some interesting details:
(i) as previously mentioned, good detection performance leads to high MOTA, to the point where a very simple IoU tracker can achieve a high MOTA of 85.5 when equipped with a strong detector.  
(ii) The same IoU tracker achieves an extremely low AP$_{50}$, as it cannot reliably output temporally consistent tracks. This shows the importance of evaluating MOT performance with both MOTA and TrackAP, as this enables deeper understanding of the actual accuracy of the model. 
(iii) Bergmann et al.~\cite{bergmann2019tracking} made the case that {\it ``a detector is all you need''} for multi-object tracking. It is clear from these results that this is unfortunately not the case. Tracking by simply regressing a bounding box over time is not sufficient to achieve good AP.
Finally, (iv) using our Siamese Track-RCNN leads to slightly higher MOTA (+2.1), but considerably better AP$_{50}$ both when using only our Track branch (+11) and both our Track and Re-id branches (+20). This is a large improvement and it further shows the importance of modelling tracking as a combination of detection, tracking and re-identification. 

\begin{table}[t]
    \footnotesize
	\begin{center}
		\begin{tabular}{l @{\hskip 1em}l @{\hskip 0.5em} | c  c | c c }
			\toprule
			Method & Tracking & MOTA$\uparrow$ & IDsw $\downarrow$ & AP$_{50}$  $\uparrow$& AP$_{75}$ $\uparrow$ \\
			\midrule
            Deepsort~\cite{Wojke2017simple} & IoU & 85.5 & 16194 & 8.07 & 2.71  \\
            Tracktor~\cite{bergmann2019tracking} & Regression & 87.6  & 12791 &  18.3 & 7.69  \\
            Siamese Track-RCNN & Siamese & 89.7 & 10700 & 29.3 & 13.4 \\
            Siamese Track-RCNN & Siamese + Re-ID & \textbf{90.2} & \textbf{6380} &  \textbf{39.7} & \textbf{18.5}  \\
			\bottomrule
		\end{tabular}
	\end{center}
	\vspace{-2mm}
	\caption{\it Results of three methods that share the same object detector, on JTA.\vspace{-5mm}}
	\label{table:results_jta}
\end{table}

\subsection{Analysis of the components of Siamese Track-RCNN}
\label{section:ablation_st_rcnn}
Here we evaluate the different branches of our Siamese Track-RCNN and quantify how much each of them contributes to the model's final performance. We present our results in table~\ref{table:ablation_jta}. All the entries in the table share the same backbone and the same Faster-RCNN-style detection branch. When we enhance this base model with our Track branch, the model achieves a TrackAP$_{50}$ of 27.8. If we further enhance the training with our Re-id branch, but avoid using it for tracking during inference (i.e., the model still relies solely on the Track branch for that), the performance improves to 29.3\%. This is interesting, as it highlights the importance of training these components jointly: adding the Re-id branch has a positive effect on the Track branch, which becomes more accurate. Moreover, when we remove the Track branch and perform tracking using the Re-id branch instead, AP$_{50}$ decreases to 26.4, which is still reasonable high. Finally, when we combine all these components in our Siamese Track-RCNN, AP$_{50}$ improves considerably to 39.7\%. This clearly shows that these components are highly complementary and are all extremely important towards accurate tracking.

\begin{figure}[t]
  \centering
  \includegraphics[width=1\textwidth]{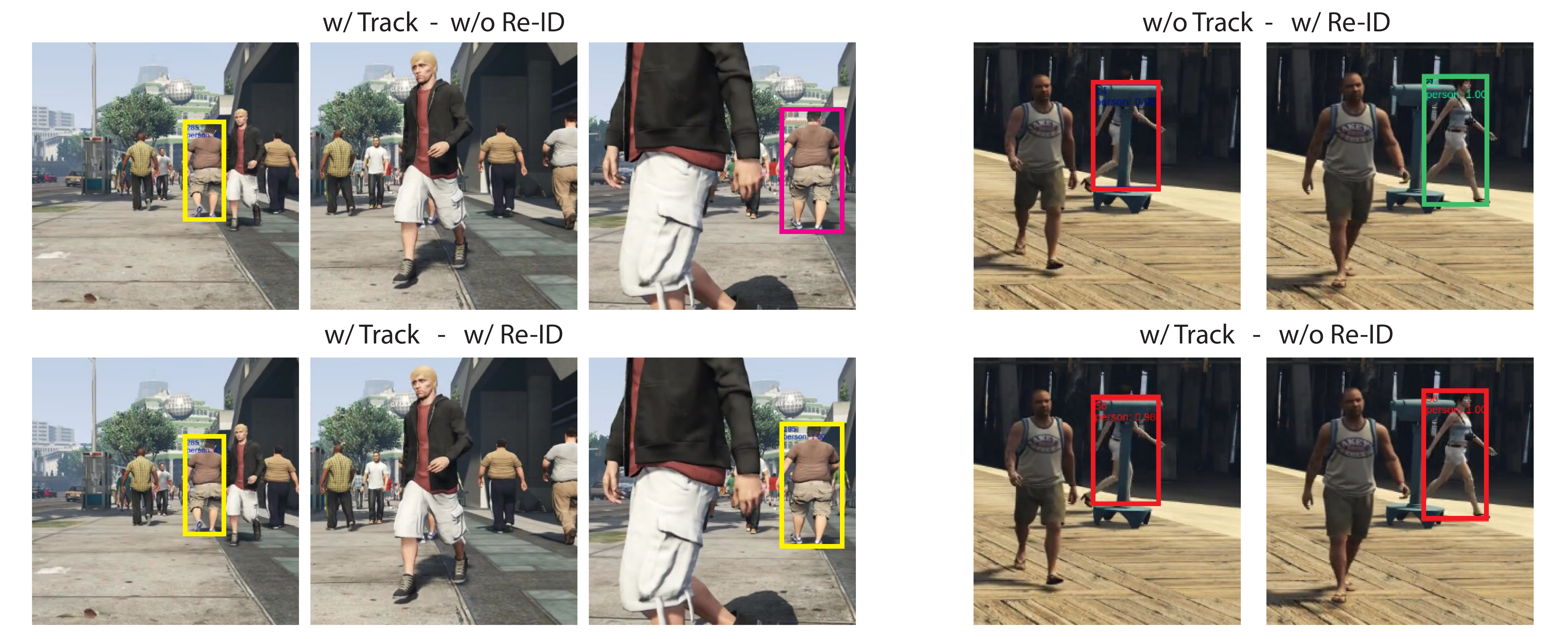}
  \vspace{-4mm}
  \caption{\it Some visual results with the different components of Siamese Track-RCNN. On the left we show the importance of our re-id branch in re-instating tracks that were terminated because of occlusion. On the right we show how our track branch is able to continue the track despite the sudden change in appearance.}
 \label{figure:reid}
\end{figure}

\begin{table}[t]
    \begin{center}
    \begin{tabular}{c @{\hskip 0.5em} c @{\hskip 0.5em} c @{\hskip 0.5em} | @{\hskip 0.5em} c @{\hskip 0.5em} c}
        \toprule
         Track & Re-id branch & Re-id branch & \multirow{2}{*}{AP$_{50}$} & \multirow{2}{*}{AP$_{75}$} \\
         branch & (training) & (inference) & & \\
         \midrule
         \checkmark & - & - & 27.8 & 12.1 \\
         \checkmark & \checkmark & - & 29.3 & 13.4 \\
         - & \checkmark & \checkmark & 26.4 & 11.2  \\
         \midrule
         \checkmark & \checkmark & \checkmark & \textbf{39.7} & \textbf{18.5}  \\
         \bottomrule
    \end{tabular}
   \end{center}
   \vspace{-2mm}
   \caption{\it \it Ablation experiments of Siamese Track-RCNN on JTA datasets. \vspace{-5mm}}
    \label{table:ablation_jta}
\end{table}

\subsection{Analysis of our parameter choices}
We now analyze some of our choices in the design of Siamese Track-RCNN.\\

\noindent {\bf Track network $\mathbb{\phi}$.}
Here we investigate the importance of using a Siamese structure in our Track branch. We compare our design against a non-Siamese network $\mathbb{\phi}$ that estimates visibility and motion of eq.~\ref{equation: siamese_track} only looking at the search region's features at time $t+\delta$: $(\hat{v}, \hat{m})= \mathbb{\phi}(\mathbf{f}_{search^{t+\delta}})$. Our model achieves considerably higher performance (29.3 vs 24.8 AP$_{50}$), validating our choice using a Siamese network as our Track branch. \\

\noindent {\bf Pairs sampling $\delta$.} 
In order to make the model more robust to different types of motions, in sec.~\ref{sec:impl_details} we proposed to augment our training set of pairs by sampling more than just two consecutive frames. Precisely, we proposed to sample frame $t$ and a second one in the range [$t+1, t+\delta$]. Here we investigate how changing the value of $\delta$, which defines the temporal sampling range, affects the overall model performance. Results are shown in table~\ref{table:delta}. Sampling frames that are too close to each ($\delta=8$) or too far away from each other ($\delta$=45) achieve the lowest performance. This is reasonable, as the former mostly includes pairs that are very similar, with can negatively affect the generalization ability of the Re-id branch; while the latter adds pairs that are too different, which makes the training of the Track branch more difficult. Instead, when we set $\delta$ to 30, which results in a temporal window of 1s on these 30fps videos, we achieve the best performance. As mentioned in sec.~\ref{sec:impl_details}, we use this value in all our results.\\

\noindent {\bf Training the model to reinstate tracks.} 
In sec.~\ref{sec:impl_details} we presented our online solver that reinstates tracks by computing re-id embedding differences and thresholding. Here we investigate how much the performance improves when we actually train a model to make this decision. We train a shallow 256-channels two-layer network for binary classification (i.e., same track or not). The input to this network is a combination of these simple features: embedding difference, mean bounding box centers, width and height, velocity, and track confidence. We experiment with two models, one that runs online and one offline. The online model operates similarly to our Siamese Track-RCNN and it computes its features by only observing a few frames of the new track. The offline model instead operates more like a postprocessor and it can observe the whole video sequence. Results are presented in table~\ref{table:ablation_solver}. The online model provides a substantial improvement in AP over just using re-id embedding and thresholding (+8.9 AP$_{50}$). This is particularly important, as this model is very lightweight and it can be easily employed in our Siamese Track-RCNN to further boost its performance. Moreover, the offline model further improves AP$_{50}$ by 1.6, showing that looking at the future can occasionally help.

% While the re-id component of the framework with a simple threshold already significantly boosts TrackAP, we also tried to compliment the network's high quality short tracks by instead using a trained solver with additional input features. We trained a small 256-channel two layer MLP network on basic bounding box and embedding features. The embedding distance from inference and track end-start gap time were input to the network. In addition, mean bounding box centers, width and height, velocities, and tracker confidences were extracted for each track, in windows of 1/4, 1/2, and 1 second from the edges of the track. The network had a single output and was trained with a BCE loss. The network was then used in both an online fashion during inference and offline as postprocessing where it was able to use larger 4 second bounding box windows and more embeddings from the new track. This basic processing provided a substantial TrackAP boost to 48.6 and 50.2 AP$_{50}$ Table~\ref{table:ablation_solver} for the online and offline methods respectively. This demonstrates that the proposed framework's high quality short tracks could benefit from the various forms of postprocessing present in the literature.

\begin{table}[t]
\footnotesize
\centering
  \begin{minipage}{.4\textwidth}
    \begin{center}
     \begin{tabular}{l @{\hskip 0.5em} | c @{\hskip 0.5em} c}
			\toprule
			Pairs sampling $\delta$ & AP$_{50}$ & AP$_{75}$ \\
			\midrule
			8 frames (0.27s)  & 31.3 & 13.8 \\
			16 frames (0.53s) & 34.7 & 15.0 \\
			30 frames (1.0s) & \bf 39.7 & \bf 18.5 \\
			45 frames (1.5s) & 30.9 & 10.9 \\
			\bottomrule
		\end{tabular}
	\end{center}
	\vspace{-2mm}
    \caption{\it Ablation study on $\delta$.}
    \label{table:delta}
  \end{minipage}
   \hspace{3em}
   \begin{minipage}{.5\textwidth}
	\begin{center}
		\begin{tabular}{l c c}
			\toprule
			Method & AP$_{50}$ & AP$_{75}$ \\
			\midrule
			Siamese Track-RCNN & 39.7 & 18.5  \\
			+ trained solver (online) & 48.6 & 25.3  \\
            + trained solver (offline) & 50.2 & 26.6  \\
			\bottomrule
		\end{tabular}
	\end{center}
	\vspace{-2mm}
	\caption{\it Trained models to reinstate tracks.}
	\label{table:ablation_solver}
\end{minipage} 
\end{table}

\section{Conclusion}

We have presented our Siamese Track-RCNN, a unified MOT framework that is built on three branches: detect, track and re-id. These branches are trained jointly in an end-to-end fashion. The online inference of Siamese Track-RCNN is efficient and accurate. We achieved the best published results on MOTChallenge (2016 and 2017 track). We conducted a thorough ablation study where we showed the importance of our components and our hyper-parameter choices. Finally, in this work we explored applying this framework to the task of person tracking, but our framework is general and extensible. In the future we will explore applying Siamese Track-RCNN to the task of general MOT across more object categories, mask tracking and pose tracking.

\clearpage
% ---- Bibliography ----
%
% BibTeX users should specify bibliography style 'splncs04'.
% References will then be sorted and formatted in the correct style.
%
\bibliographystyle{splncs04}
\bibliography{eccv2020_bib}

\clearpage
\title{Appendix}
\author{}
\institute{}
\maketitle

\section{Visual examples}

\begin{figure}[t]
    \centering
    \includegraphics[width=1.0\textwidth]{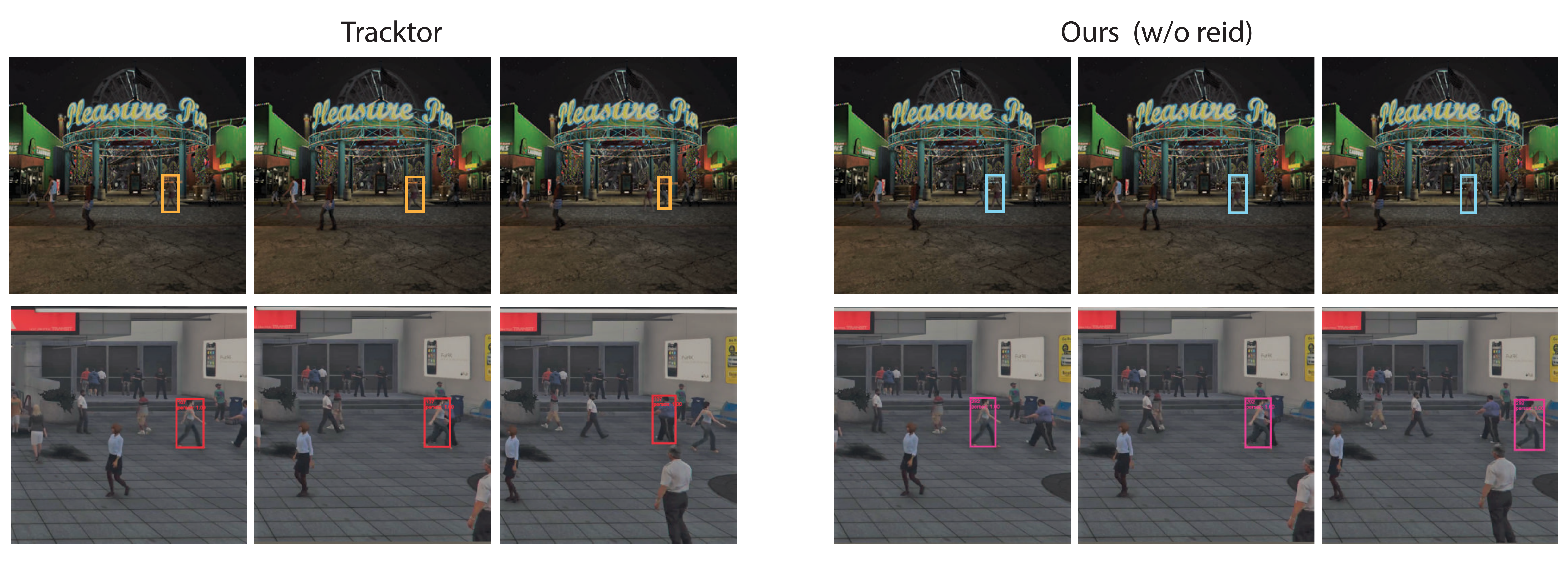}
    \caption{In the highlighted examples, Tracktor \cite{bergmann2019tracking} jumps to different tracks when two people are crossing each other, whereas Siamese Track-RCNN is capable of following the right person.}
    \label{figure:sumpplement_tracktor_vs_track}
\end{figure}

In Fig. \ref{figure:sumpplement_tracktor_vs_track}, we show the tracking results of Tracktor \cite{bergmann2019tracking} and Siamese Track-RCNN (w/o re-id branch).

\begin{figure}[t]
    \centering
    \includegraphics[width=1.0\textwidth]{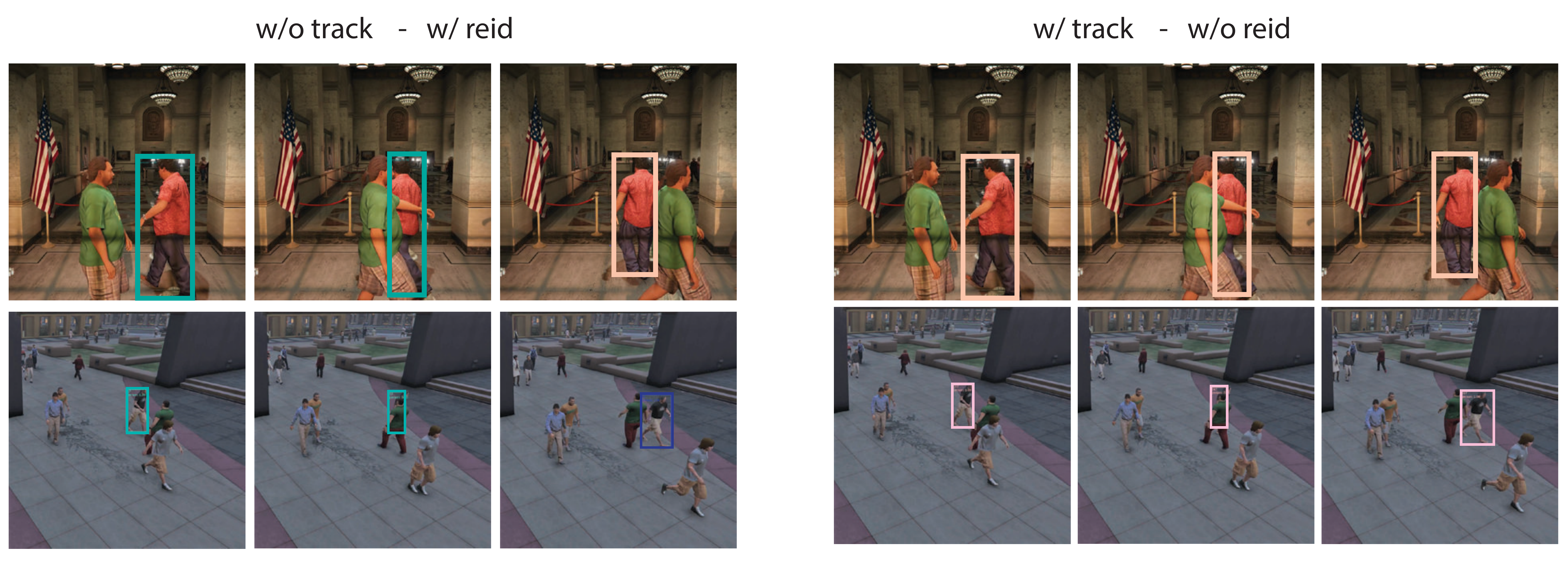}
    \caption{In the highlighted examples, tracking with outputs of re-id branch (i.e. embedding) alone leads to split tracklets when partial occlusion happens, whereas the Siamese track branch is able to track through those cases.  }
    \label{figure:sumpplement_embedding_vs_track}
\end{figure}

\noindent In Fig. \ref{figure:sumpplement_embedding_vs_track}, we show the tracking results of online Siamese Track-RCNN with track branch or with re-id branch only.

\begin{figure}[t]
    \centering
    \includegraphics[width=1.0\textwidth]{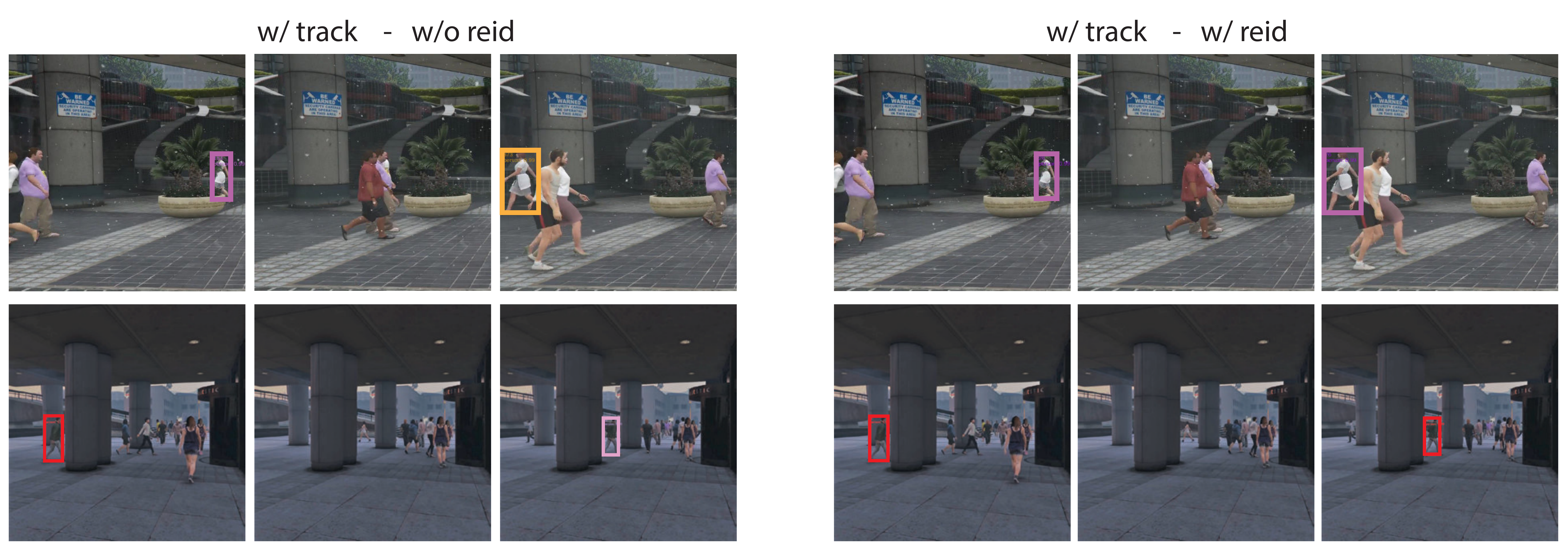}
    \caption{The highlighted examples demonstrate the significance of re-id branch for Siamese Track-RCNN to track through full occlusion. }
    \label{figure:sumpplement_embedding}
\end{figure}

\noindent In Fig. \ref{figure:sumpplement_embedding}, we show the tracking results of online Siamese Track-RCNN with and without incorporating the outputs of re-id branch during inference.

\begin{figure}[t]
    \centering
    \includegraphics[width=1.0\textwidth]{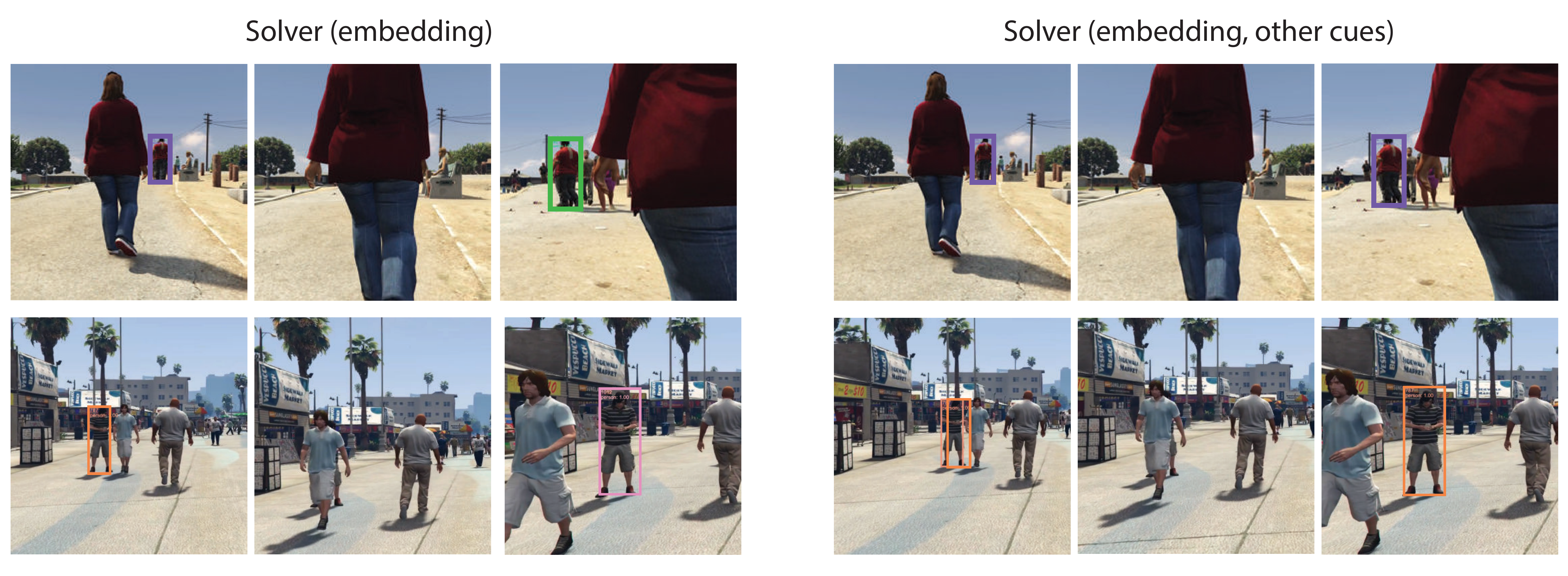}
    \caption{The highlighted examples illustrate the importance of  using richer cues (i.e. embedding, velocity, etc.) to reinstate the tracks especially when their appearance has largely changed (e.g. due to long-time occlusions).}
    \label{figure:sumpplement_learned_solver}
\end{figure}

\noindent In Fig. \ref{figure:sumpplement_learned_solver}, we show the tracking results of two versions of online Siamese Track-RCNN. One only incorporates the outputs of re-id branch (embedding) to \textbf{reinstate} tracks, and the other one incorporates both embedding and other cues (e.g. bounding box velocity, sizes, etc.) to \textbf{reinstate} tracks.

\section{Detailed results on MOTChallenge}
In Table \ref{table:supplement_results_mot16} and \ref{table:supplement_results_mot17}, we present the sequence-level result summary of MOTChallenge of 2016 and 2017 respectively. 
\begin{table}[]
    \begin{center}
    \begin{tabular}{l @{\hskip 1em} l @{\hskip 0.5em} c  c  c @{\hskip 0.5em} c @{\hskip 0.5em} l @{\hskip 0.5em} l @{\hskip 1em} l}
    \toprule
    Sequence & Detection & MOTA ($\uparrow$) & IDF1 ($\uparrow$) & MT ($\uparrow$) & ML ($\downarrow$) & FP ($\downarrow$) & FN ($\downarrow$) & IDsw ($\downarrow$) \\
    \midrule
    MOT16-01 & DPM \cite{dpm} & 45.1	& 47.5 & 21.7\% & 39.1\% & 12 & 3485 & 11 \\
    MOT16-03 & DPM & 72.3	& 70.1 & 47.3\% & 10.8\% & 3031 & 25762 & 122 \\
    MOT16-06 & DPM & 52.8	& 52.8 & 20.4\%	& 40.3\% & 358	& 4983	& 106 \\
    MOT16-07 & DPM & 51.7	& 49.7 & 18.5\%	& 18.5\% & 206	& 7588	& 90 \\
    MOT16-08 & DPM & 34.7	& 36.4 & 12.7\%	& 36.5\% & 246	& 10612	& 64 \\
    MOT16-12 & DPM & 49.0	& 55.9 & 20.9\%	& 45.3\% & 194  & 4018	& 21 \\
    MOT16-14 & DPM & 32.8	& 38.5 & 6.70\%	& 46.3\% & 342	& 11928	& 142 \\
    \midrule
    \multicolumn{2}{c}{All} & 59.8 & 60.8 & 22.0\% & 34.5\% & 4389 & 68376 & 556 \\
    \bottomrule
    \end{tabular}
    \end{center}
    \caption{Detailed result summary on MOT16 test videos.}
    \label{table:supplement_results_mot16}
\end{table}

\begin{table}[t]
   \footnotesize
    \begin{center}
    \begin{tabular}{l @{\hskip 1em} l c  c  c @{\hskip 0.5em} c @{\hskip 0.5em} l @{\hskip 0.5em} l @{\hskip 0.5em} l}
    \toprule
    Sequence & Detection & MOTA ($\uparrow$) & IDF1 ($\uparrow$) & MT ($\uparrow$) & ML ($\downarrow$) & FP ($\downarrow$) & FN ($\downarrow$) & IDsw ($\downarrow$) \\
    \midrule
    MOT17-01 & DPM \cite{dpm} & 44.8 & 47.2 & 20.8\% & 41.7\% & 12 & 3540 & 11 \\
    MOT17-03 & DPM & 72.3	& 70.0 & 47.3\% & 10.8\% & 3009 & 25859 & 120 \\
    MOT17-06 & DPM & 52.3	& 52.1 & 19.4\%	& 40.5\% & 320	& 5193	& 106 \\
    MOT17-07 & DPM & 49.9	& 48.5 & 16.7\%	& 30.0\% & 209	& 8161	& 93 \\
    MOT17-08 & DPM & 27.7	& 30.7 & 10.5\%	& 47.4\% & 226	& 14979	& 64 \\
    MOT17-12 & DPM & 47.2	& 54.5 & 19.8\%	& 47.3\% & 178  & 4380	& 21 \\
    MOT17-14 & DPM & 32.8	& 38.5 & 6.70\%	& 46.3\% & 342	& 11928	& 142 \\
    \midrule
     MOT17-01 & FRCNN \cite{ren2015faster} & 46.1	& 48.2 & 20.8\% & 41.7\% & 34 & 3434 & 10 \\
    MOT17-03 & FRCNN & 72.0	& 68.0 & 46.6\% & 12.8\% & 3169 & 26072 & 108 \\
    MOT17-06 & FRCNN & 56.8	& 54.5 & 25.7\%	& 28.4\% & 442	& 4482	& 171 \\
    MOT17-07 & FRCNN & 47.8	& 48.6 & 18.3\%	& 28.3\% & 306	& 8422	& 90 \\
    MOT17-08 & FRCNN & 27.0	& 31.6 & 11.8\%	& 50.0\% & 239	& 15120	& 53 \\
    MOT17-12 & FRCNN & 43.2	& 55.6 & 16.5\%	& 50.5\% & 268  & 4637	& 17 \\
    MOT17-14 & FRCNN & 35.2	& 39.1 & 8.50\%	& 43.3\% & 544	& 11178	& 253 \\
    \midrule
     MOT17-01 & SDP \cite{yang2016exploit} & 48.4	& 48.9 & 20.8\% & 37.5\% & 51 & 3262 & 15 \\
    MOT17-03 & SDP & 78.4	& 72.3 & 61.5\% & 8.10\% & 4022 & 18442 & 141 \\
    MOT17-06 & SDP & 57.0	& 55.1 & 29.7\%	& 28.8\% & 534	& 4358	& 177 \\
    MOT17-07 & SDP & 52.8	& 49.6 & 23.3\%	& 26.7\% & 341	& 7541	& 99 \\
    MOT17-08 & SDP & 28.2	& 33.3 & 14.5\%	& 44.7\% & 338	& 14756	& 83 \\
    MOT17-12 & SDP & 46.6	& 54.6 & 19.8\%	& 48.4\% & 284  & 4325	& 19 \\
    MOT17-14 & SDP & 38.4	& 41.4 & 7.30\%	& 40.2\% & 664	& 10450	& 275 \\
    \midrule
    \multicolumn{2}{c}{All} & 59.6 & 60.1 & 23.9\% & 33.9\% & 15532 & 210519 & 2068 \\
    \bottomrule
    \end{tabular}
    \end{center}
    \caption{Detailed result summary on MOT17 test videos.}
    \label{table:supplement_results_mot17}
\end{table}

\end{document}